\documentclass[runningheads]{llncs}

\usepackage{eccv}

\usepackage{multirow}
\usepackage{wrapfig}

\usepackage{eccvabbrv}

\usepackage{graphicx}
\usepackage{booktabs}

\usepackage[accsupp]{axessibility}  

\usepackage{hyperref}

\usepackage{orcidlink}

\usepackage[misc]{ifsym}

\begin{document}

\title{Hierarchical Spatial and Channel Aggregation for Cross-domain Few-shot Segmentation} 

\titlerunning{Hierarchical Spatial and Channel Aggregation for CD-FSS}

\author{Sujun Sun\inst{1,2}\orcidlink{0009-0002-0647-1148} \and
Mingwu Ren\inst{1,2}\orcidlink{0000-0001-5576-3281} \and
Haofeng Zhang\inst{1,2}\textsuperscript{(\Letter)}\orcidlink{0000-0002-4039-7618}}

\authorrunning{S. Sun et al.}

\institute{School of Computer Science and Engineering, Nanjing University of Science and Technology, China \and
State Key Laboratory of Intelligent Manufacturing of Advanced Construction Machinery, China\\
\email{\{egg, renmingwu, zhanghf\}@njust.edu.cn}}

\maketitle

\begin{abstract}
Cross-domain Few-shot Segmentation (CD-FSS) aims to le-arn generalizable segmentation capability from abundant annotated samples in the source domain, enabling accurate segmentation of novel classes in the target domain with only a few annotated samples. Existing CD-FSS methods mainly focus on mitigating feature distribution shifts caused by style gaps while ignoring significant differences in class semantic granularity and discriminative attributes across domains, leading to two key degradations in support–query matching: semantic over-alignment and attribute over-alignment. To this end, we propose the Dual Hierarchical Aggregation Network (DHANet), which comprises three key modules. First, the Hierarchical Spatial Aggregation (HSA) module performs multi-scale region aggregation of pixel features along the spatial dimension, generating hierarchical semantic-enhanced features to alleviate semantic over-alignment. Additionally, the HCA module conducts multi-scale attribute aggregation along the channel dimension, generating hierarchical attribute-enhanced features to mitigate attribute over-alignment. Finally, we propose the Online Probabilistic Semantic Bank (OPSB), which progressively constructs and updates class probability distributions from query predictions during inference, and samples multiple pseudo-prototypes as additional support information to mitigate insufficient support. Extensive experiments on four target-domain datasets demonstrate that our method achieves state-of-the-art performance.
\end{abstract}

\begin{figure}[tb]
    \centering
    \includegraphics[width=0.98\linewidth]{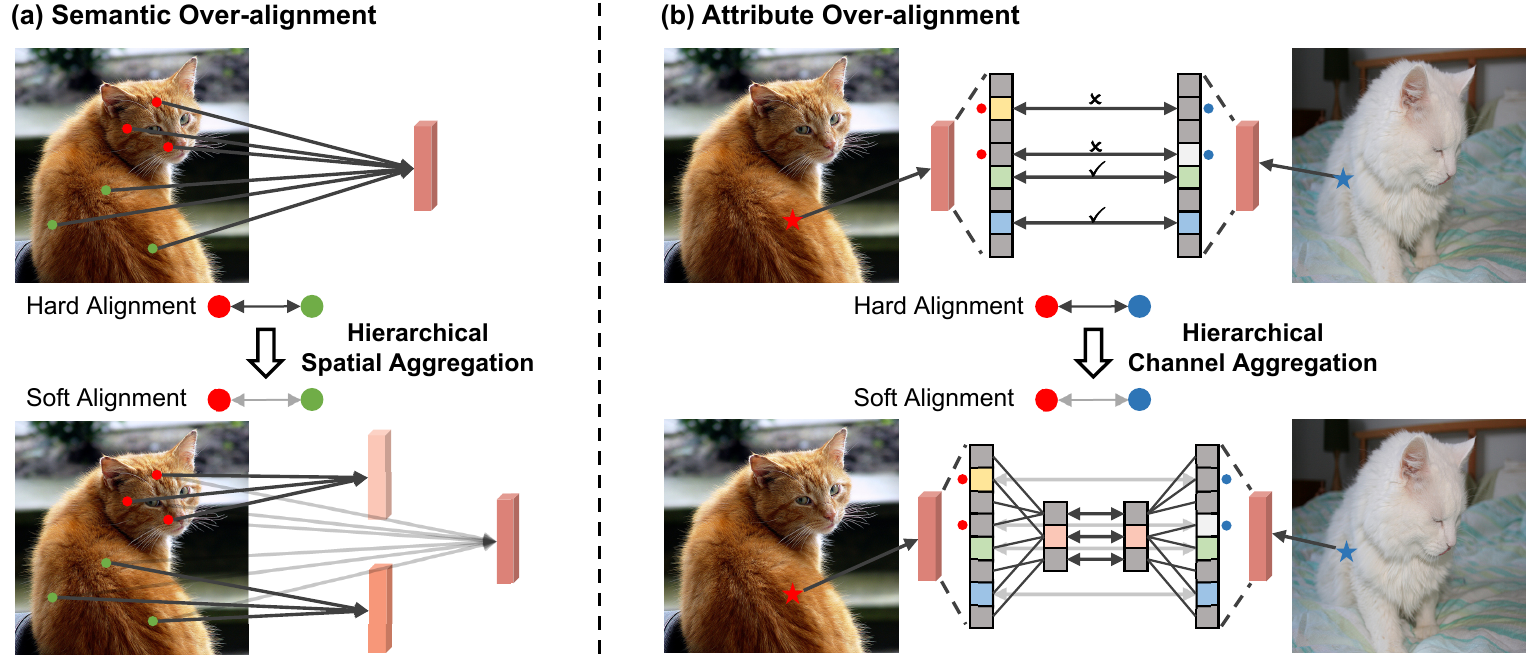}
    \caption{Motivation of DHANet. (a) Semantic over-alignment prevents the model from distinguishing semantic categories at other segmentation granularities (\eg, a cat’s head and body). Hierarchical spatial aggregation aligns features at multiple granularities, adapting to target domains with varying granularities. (b) Attribute over-alignment causes the degradation of source-insensitive attributes. Hierarchical channel aggregation mitigates hard channel-wise alignment, preserving sensitivity to diverse attributes.}
    \vspace{-3ex}
    \label{fig:motivation}
\end{figure}

\section{Introduction}
\label{sec:intro}

Driven by robust network designs \cite{guo2022segnext, yan2024multi, fu2025segman} and large-scale densely annotated datasets \cite{liao2022kitti, zhou2017scene}, semantic segmentation has achieved remarkable progress in recent years. However, when confronted with novel classes that have only a few annotated samples, the performance of these models often degrades sharply, failing to meet the demand for rapid deployment in real-world scenarios. To alleviate the reliance on large-scale annotated data, Few-shot Segmentation (FSS) has been proposed \cite{fan2022self, sun2024vrp, xu2025unlocking, min2021hypercorrelation}. It leverages meta-learning to learn transferable category correspondence on base classes, enabling the segmentation of novel classes using only a few annotated samples. Conventional FSS methods inherently assume that base and novel classes share the same data domain. However, in realistic scenarios, substantial domain gaps typically exist between the source and target domains, leading to significant performance drops of existing FSS models. This practical challenge has motivated Cross-domain Few-shot Segmentation (CD-FSS) \cite{lei2022cross}, which has attracted increasing attention.

Existing CD-FSS methods typically focus on mitigating cross-domain distribution shifts caused by style gaps. Prevalent techniques include performing style perturbations \cite{su2024domain, liu2025devil}, extracting domain-agnostic features \cite{lei2022cross, he2024apseg, tong2025adapter}, or fine-tuning with limited target domain annotated data \cite{peng2025sam, nie2024cross}. Although effective in reducing global feature distribution gaps between source and target domains, these strategies are ignoring substantial category differences across domains, which leads to support–query over-alignment during training, including semantic over-alignment and attribute over-alignment.

As shown in \cref{fig:motivation}(a), the training objective forces all pixel features within the same source domain class to align with the class prototype, causing the model to only excel at distinguishing foreground and background at the same segmentation granularity as the source domain (\eg, cat and other classes), while failing to discriminate semantic objects at other granularities (\eg, a cat’s head and body), \ie, semantic over-alignment in the source domain. When the target domain requires a finer or coarser segmentation granularity, the model’s foreground–background discrimination capability degrades significantly. As shown in \cref{fig:motivation}(b), different instances of the same class often exhibit attribute variations (\eg, white and orange cats differ in the color attribute). These source-insensitive attributes may be activated by different channels in support and query features, and the incorrect matching caused by channel-wise alignment forces these channels to gradually degenerate, rendering them unable to capture such source-insensitive attributes, \ie, attribute over-alignment in the source domain. When the target domain is sensitive to these attributes (\eg, dermoscopic images \cite{codella2019skin} rely on color to distinguish lesions), the consistency of intra-class features across images weakens, substantially degrading segmentation performance.

Although some recent works \cite{tong2025self, sun2026bridging} have started to address semantic over-alignment, they typically rely on external models \cite{xu2024learning} or exhibit limited adaptability to target domains with varying segmentation granularities, while also ignoring the attribute over-alignment. To simultaneously alleviate semantic and attribute over-alignment, we propose the Dual Hierarchical Aggregation Network (DHANet). The core idea is to perform hierarchical aggregation along the spatial and channel dimensions, respectively, to capture semantic and attribute representations that are discriminative at different granularities. Such hierarchical representations not only convert hard alignment at a single granularity into soft alignment during training, mitigating over-alignment, but also provide multi-granularity information during testing, adapting to differences in class semantic granularity and discriminative attributes across different domains.

Specifically, our DHANet consists of three key modules. First, the Hierarchical Spatial Aggregation (HSA) module performs multi-scale region aggregation of pixel features along the spatial dimension via multiple groups of spatial slots and slot attention, producing hierarchical region prototypes. These prototypes are then used to enhance original features, generating hierarchical semantic-enhanced features. Second, the Hierarchical Channel Aggregation (HCA) module treats channels as aggregation primitives and performs aggregation along the channel dimension with multi-level channel slots, forming hierarchical attribute prototypes. These prototypes are then concatenated with original features to produce hierarchical attribute-enhanced representations. To prevent slot homogenization, we impose orthogonal constraints on the spatial and channel slots at each level. Finally, to further alleviate insufficient support information at test time, we propose an Online Probabilistic Semantic Bank (OPSB) module, which models class prototypes as probability distributions. It constructs and updates class probabilistic prototypes using high-confidence features extracted from the query predictions of each episode, and samples multiple pseudo-prototypes from the latest distributions as additional support information.

In summary, our contributions are as follows:
\begin{itemize}
\item We are the first to simultaneously focus on semantic over-alignment and attribute over-alignment in CD-FSS, and propose the HSA and HCA modules to aggregate discriminative semantic and attribute representations at different granularities along the spatial and channel dimensions, respectively.
\item We propose an OPSB module to alleviate insufficient support information at test time, which constructs and updates class probabilistic prototypes using high-confidence query predictions and samples multiple pseudo-prototypes to provide additional support.
\item Extensive experiments on four standard target-domain datasets demonstrate that our method achieves state-of-the-art performance on CD-FSS.
\end{itemize}

\section{Related Works}
\label{sec:related}

\subsection{Few-shot Segmentation}
FSS aims to predict dense masks for novel classes in query images using only a few annotated support images. Early FSS methods \cite{fan2022self, wang2019panet, yang2020prototype, liu2020part} typically summarize the support set into class-representative prototypes and perform segmentation by measuring the similarity between query features and these prototypes. To address the loss of spatial details inherent in prototype computation, matching-based methods \cite{peng2023hierarchical, xu2023self, xu2024eliminating, min2021hypercorrelation} fully exploit support information by computing dense pixel-to-pixel correlations between support and query. Recently, several methods \cite{wang2024rethinking, leng2025multi, bi2024prompt} introduce textual semantic features as additional support to alleviate the insufficiency of support information caused by large intra-class variance. Meanwhile, some methods \cite{sun2024vrp, zhang2024bridge, xu2025unlocking} leverage the large-scale pretraining and generalization capabilities of vision foundation models \cite{oquab2023dinov2, kirillov2023segment} to enhance FSS, achieving remarkable performance gains. However, these methods are not designed with cross-domain transfer in mind, resulting in limited generalization to novel domains.

\subsection{Cross-domain Few-shot Segmentation}
The limitations of FSS methods in cross-domain scenarios have motivated increasing attention to CD-FSS, which requires models trained on the source domain to generalize to novel classes across diverse domains. Some methods \cite{su2024domain, liu2025devil} apply style perturbations to features during source training to simulate cross-domain style variations, thereby narrowing the feature distribution gaps between source and diverse target domains. Others \cite{lei2022cross, he2024apseg, tong2025adapter} aim to decompose or transform features into domain-agnostic spaces, generating domain-agnostic representations with cross-domain generalizability. Furthermore, most methods \cite{peng2025sam, nie2024cross, herzog2024adapt, tong2024lightweight} leverage limited target annotated samples for fine-tuning to better adapt to the target domain. However, these methods ignore the substantial differences in semantic granularity and discriminative attributes across domains, leading to semantic over-alignment and attribute over-alignment. SDRC \cite{tong2025self} decomposes features into representations focusing on distinct semantic regions to alleviate semantic over-alignment, but this single-granularity decoupling exhibits insufficient adaptability to different domains. Similarly, HSL \cite{sun2026bridging} learns hierarchical semantic features with the assistance of multi-scale superpixel segmentation priors, yet its reliance on superpixel segmentation models limits its applicability. In contrast, we adaptively construct hierarchical semantic and attribute structures to simultaneously address both semantic and attribute over-alignment.

\begin{figure}[t]
\centering
    \includegraphics[width=0.98\linewidth]{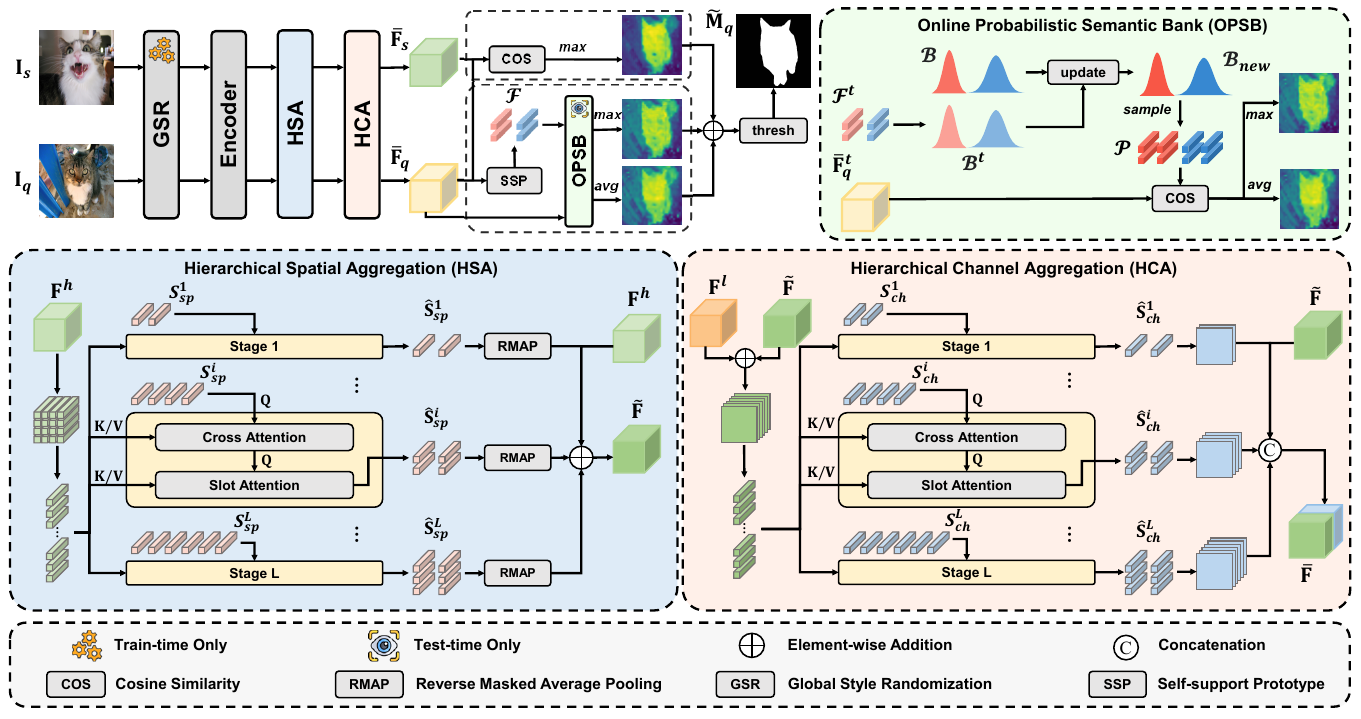}
    \caption{Overview of our method. The support and query images are first fed into GSR for style perturbation, and features are extracted by the image encoder. Subsequently, the HSA and HCA modules sequentially perform hierarchical semantic and attribute enhancement on the features. The enhanced features are then fed into the main branch and the auxiliary branch to compute the query foreground confidence maps, respectively. Finally, the foreground confidence maps from two branches are fused and thresholded to obtain the query prediction.}
    \vspace{-2ex}
    \label{fig:framework}
\end{figure}

\section{Methodology}
\label{sec:method}

\subsection{Problem Definition}
CD-FSS aims to transfer a model trained on the source domain $\mathcal{D}_s=\{X_s,Y_s\}$ to the target domain $\mathcal{D}_t = \{X_t, Y_t\}$, where the source and target domains are disjoint in both the data distribution $X$ and class labels $Y$, \ie, $X_s \neq X_t$ and $Y_s \cap Y_t = \emptyset$. CD-FSS models are typically trained using the episode-based meta-learning paradigm, where the training and testing data consist of multiple episodes randomly sampled from $\mathcal{D}_s$ and $\mathcal{D}_t$, respectively. Each episode contains a support set $\mathcal{S} = \{(\mathbf{I}^i_s, \mathbf{M}^i_s)\}_{i=1}^K$ and a query set $\mathcal{Q} = \{(\mathbf{I}_q, \mathbf{M}_q)\}$, where $K$ denotes the number of support samples, $\mathbf{I}_s$ and $\mathbf{I}_q$ represent the support and query images, and $\mathbf{M}_s$ and $\mathbf{M}_q$ are their corresponding binary masks. The model extracts support information from the support set $\mathcal{S}$ to predict the segmentation mask for the query image $\mathbf{I}_q$.

\subsection{Overview}
To alleviate both semantic over-alignment and attribute over-alignment during training, as well as insufficient support information during testing, we propose the Dual Hierarchical Aggregation Network (DHANet), whose overall architecture is illustrated in \cref{fig:framework}. 

Given support and query images, we first apply Global Style Randomization (GSR) to the input images using random convolution, following \cite{liu2025devil, sun2026bridging}. The perturbed images are then fed into a weight-shared encoder to extract the original support and query features. Next, the features are sequentially passed through Hierarchical Spatial Aggregation (HSA) and Hierarchical Channel Aggregation (HCA) modules to obtain features with hierarchical enhancement in both semantics and attributes. In the main branch, the initial foreground confidence map is obtained through dense support-query feature matching. In the auxiliary branch, the Online Probabilistic Semantic Bank (OPSB) module updates class probability distributions and samples pseudo-prototypes from them to provide additional support information for computing an auxiliary foreground confidence map. Finally, the confidence maps from two branches are combined, and the final query prediction is obtained by adaptive thresholding in \cite{sun2026bridging}. Note that GSR is used only during training, and the auxiliary branch is used only during testing.

\subsection{Hierarchical Spatial Aggregation}
The core idea of HSA is to generate hierarchical semantic-enhanced features that are inter-class discriminative and intra-class consistent at different semantic granularities. During source domain training, these hierarchical representations enable support and query to align at multiple semantic granularities, thereby mitigating over-alignment at a single granularity. During target domain testing, these representations enhance regional consistency at different scales, preventing interference from noise. Motivated by slot attention \cite{locatello2020object, ma2023attrseg}, which aggregates features from specific image regions via a zero-sum game mechanism for unsupervised region partitioning, we introduce a hierarchical spatial aggregation process. Specifically, we employ varying numbers of slots at each level to aggregate semantic features at diverse granularities.

Specifically, HSA is designed as a hierarchical architecture with $L_s$ stages. At the $i$-th stage, we allocate a set of learnable spatial slots $\mathbf{S}^i_{sp} \in \mathbb{R}^{N^i_{sp} \times C}$, where $N^i_{sp}$ denotes the number of spatial slots at this stage and $C$ is the channel depth. Given the deep high-level feature $\mathbf{F}^{h} \in \mathbb{R}^{H \times W \times C}$ extracted by the image encoder, where $H$ and $W$ denote the feature height and width, respectively, we flatten the spatial dimensions to obtain $\mathbf{F}^{h} \in \mathbb{R}^{HW \times C}$. First, we aggregate initial information for the slots via cross-attention, employing $\mathbf{S}^i_{sp}$ as queries and $\mathbf{F}^{h}$ as keys and values:
\begin{equation}
\mathbf{\tilde{S}}^i_{sp} = \operatorname{CrossAttn}(\mathbf{S}^i_{sp}, \mathbf{F}^h),
\label{eq:cross_attn}
\end{equation}
where $\mathbf{\tilde{S}}^i_{sp}$ denotes the enhanced spatial slots. Then, $\mathbf{\tilde{S}}^i_{sp}$ is fed into slot attention for iterative updating:
\begin{equation}
\mathbf{\hat{S}}^i_{sp},\mathbf{\hat{A}}^i_{sp}=\operatorname{SlotAttn}(\mathbf{\tilde{S}}^i_{sp},\mathbf{F}^h),
\label{eq:slot_attn}
\end{equation}
where $\mathbf{\hat{S}}^i_{sp} \in \mathbb{R}^{N^i_{sp} \times C}$ represents region prototypes aggregated from different spatial regions of $\mathbf{F}^{h}$, and $\mathbf{\hat{A}}^i_{sp} \in \mathbb{R}^{N^i_{sp} \times HW}$ denotes the spatial attention maps between $\mathbf{\hat{S}}^i_{sp}$ and $\mathbf{F}^{h}$. To obtain the spatial region corresponding to each region prototype, we determine the index of the maximum value of $\mathbf{\hat{A}}^i_{sp}$ along the slot dimension $N^i_{sp}$, yielding the spatial assignment masks $\mathbf{M}^i_{sp} \in \mathbb{R}^{N^i_{sp} \times H \times W}$ for each region prototype.

To apply each region prototype to its associated pixels, we employ the reverse process of masked average pooling (RMAP) to obtain a region feature map $\mathbf{F}^i$. This process is formulated as:
\begin{equation}
\mathbf{F}^i(x,y)={\sum}^{N^i_{sp}}_{j}\mathbf{\hat{S}}^{ij}_{sp}\mathbf{M}^{ij}_{sp}(x,y),
\end{equation}
where $(x,y)$ denotes the spatial coordinate, $\mathbf{\hat{S}}^{ij}_{sp}\in \mathbb{R}^{C}$ is the $j$-th region prototype at stage $i$, and $\mathbf{M}^{ij}_{sp}$ is its corresponding spatial assignment mask.

Finally, we enhance $\mathbf{F}^h$ with region feature maps from all stages to obtain the hierarchical semantic-enhanced feature $\mathbf{\tilde{F}}$:
\begin{equation}
\mathbf{\tilde{F}}=\mathbf{F}^h+\frac{1}{L_s}{\sum}_{i}^{L_s}\mathbf{F}^i.
\end{equation}

\subsection{Hierarchical Channel Aggregation}
Intra-class instances may exhibit attribute variations, meaning the class is insensitive to certain attributes. As a result, the same semantic region may be activated by different channels across support and query features, causing incorrect matching under channel-wise alignment. To address this, we propose an HCA module, which performs hierarchical aggregation along the channel dimension to group channels that activate the same region into hierarchical attribute prototypes. These prototypes are then used for support–query alignment, mitigating the adverse effect of hard channel-wise matching.

HCA follows a process similar to HSA. It is also designed as a hierarchical architecture with $L_c$ stages, while the aggregation is performed along the channel dimension rather than the spatial dimension. Specifically, at the $i$-th stage, we allocate a set of learnable channel slots $\mathbf{S}^i_{ch} \in \mathbb{R}^{N^i_{ch} \times HW}$, where $N^i_{ch}$ is the number of channel slots at this stage. Given the feature $\mathbf{\tilde{F}}$, we first combine it with the shallow feature $\mathbf{F}^{l} \in \mathbb{R}^{H \times W \times C}$ to produce a more detail-rich feature $\mathbf{\hat{F}}=\mathbf{\tilde{F}}+\mathbf{F}^{l}$. We then flatten its spatial dimensions and transpose it to obtain $\mathbf{\hat{F}} \in \mathbb{R}^{C \times HW}$. Following the same steps as \cref{eq:cross_attn} and \cref{eq:slot_attn}, we obtain attribute prototypes $\mathbf{\hat{S}}^i_{ch} \in \mathbb{R}^{N^i_{ch} \times HW}$ aggregated from the channels of $\mathbf{\hat{F}}$:
\begin{equation}
\mathbf{\tilde{S}}^i_{ch} = \operatorname{CrossAttn}(\mathbf{S}^i_{ch}, \mathbf{\hat{F}}), \quad
\mathbf{\hat{S}}^i_{ch},\mathbf{\hat{A}}^i_{ch}=\operatorname{SlotAttn}(\mathbf{\tilde{S}}^i_{ch},\mathbf{\hat{F}}).
\end{equation}

To avoid undermining the attribute diversity of features, we do not apply attribute prototypes to their corresponding channels using the region-filling strategy in HSA. In contrast, we concatenate $\mathbf{\hat{S}}^i_{ch}$ from all stages with the feature $\mathbf{\tilde{F}}$ along the channel dimension to obtain the hierarchical attribute-enhanced feature $\mathbf{\bar{F}}$:
\begin{equation}
\mathbf{\bar{F}}=
\begin{bmatrix}
\mathbf{\tilde{F}};\mathbf{\hat{S}}_{ch}^1;\mathbf{\hat{S}}_{ch}^2;\ldots;\mathbf{\hat{S}}_{ch}^{L_c}
\end{bmatrix}\in\mathbb{R}^{C^{\prime}\times HW}, \quad C^{\prime} = C+{\sum}_{i=1}^{L_c}N_{ch}^i,
\end{equation}
where $[\cdot; \cdot]$ denotes concatenation along the channel dimension. Finally, we restore the spatial dimensions of $\mathbf{\bar{F}}$ and transpose it to obtain the final hierarchical attribute-enhanced feature $\mathbf{\bar{F}} \in \mathbb{R}^{H \times W \times C^{\prime}}$.

\subsection{Online Probabilistic Semantic Bank}
The limited labeled support data available at test time offers restricted class information. Some works \cite{lei2022cross, chen2024cross} perform continuous fine-tuning using query predictions from each episode during testing to exploit additional class information in query images. However, such continuous fine-tuning entails the risk of knowledge forgetting. Instead, we avoid per-episode fine-tuning and dynamically accumulate reliable class priors to construct a semantic bank. An intuitive method is to progressively maintain and update a single class prototype using query predictions. However, for target domain classes with large intra-class variations, query features may still exhibit a considerable distance from the accurate prototype. We overcome this limitation by establishing a probabilistic semantic bank.

Specifically, we model class prototypes as Gaussian distributions. Given a set of high-confidence features $\mathcal{F}=\{\mathbf{f}^i\}_{i=1}^{N}$, we estimate distribution parameters as:
\begin{equation}
\boldsymbol{\mu}=\frac{1}{N}{\sum}_{i=1}^{N}\mathbf{f}^{i},\quad \boldsymbol{\sigma}^{2}=\frac{1}{N-1}{\sum}_{i=1}^{N}(\mathbf{f}^{i}-\boldsymbol{\mu})^{2},
\label{eq:gaussian}
\end{equation}
where $\boldsymbol{\mu}$ and $\boldsymbol{\sigma}^{2}$ denote the mean and variance, respectively, and $N$ represents the number of samples.

During testing, we maintain an online semantic bank for each class:
\begin{equation}
\mathcal{B} = \{\{\boldsymbol{\mu}_{fg}, \boldsymbol{\sigma}^2_{fg}, N_{fg}\}, \{\boldsymbol{\mu}_{bg}, \boldsymbol{\sigma}^2_{bg}, N_{bg}\}\}.
\end{equation}
For the $t$-th episode, we first obtain the support feature $\mathbf{\bar{F}}^t_s$ and query feature $\mathbf{\bar{F}}^t_q$. High-confidence query foreground feature set $\mathcal{F}^t_{fg}$ and background feature set $\mathcal{F}^t_{bg}$ are then extracted via SSP \cite{fan2022self}. Subsequently, we estimate the episode-specific foreground distribution $\{\boldsymbol{\mu}_{fg}^t, (\boldsymbol{\sigma}^{t}_{fg})^2, N_{fg}^t\}$ using \cref{eq:gaussian} and fuse it with the existing statistics in the semantic bank:
\begin{equation}
\begin{aligned}
N_{new}&=N_{fg}+N_{fg}^t, \quad \boldsymbol{\mu}_{new}=\frac{N_{fg}\boldsymbol{\mu}_{fg}+N_{fg}^t\boldsymbol{\mu}_{fg}^t}{N_{new}}, \\
\boldsymbol{\sigma}^2_{new}&=\frac{N_{fg}(\boldsymbol{\sigma}^2_{fg}+(\boldsymbol{\mu}_{fg}-\boldsymbol{\mu}_{new})^2)+N_{fg}^t((\boldsymbol{\sigma}^{t}_{fg})^2+(\boldsymbol{\mu}_{fg}^t-\boldsymbol{\mu}_{new})^2)}{N_{new}}.
\end{aligned}
\end{equation}
Finally, the semantic bank is updated with the fused statistics:
\begin{equation}
\{\boldsymbol{\mu}_{fg}, \boldsymbol{\sigma}^2_{fg}, N_{fg}\} \leftarrow \{\boldsymbol{\mu}_{new}, \boldsymbol{\sigma}^2_{new}, N_{new}\}.
\end{equation}
The background distribution is updated in the same manner.

\subsection{Segmentation}
During testing, given the query feature $\mathbf{\bar{F}}_q$, support feature $\mathbf{\bar{F}}_s$, and support mask $\mathbf{M}_s$, we compute the query foreground confidence map via two branches.

In the main branch, we compute the cosine similarity between $\mathbf{\bar{F}}_q$ and all foreground support features, and take the maximum along the support dimension to obtain the foreground similarity map $\mathbf{M}_{fg}$:
\begin{equation}
\mathbf{M}_{fg}(\mathbf{u}_q)=\max_{\mathbf{u}_s} \cos(\mathbf{\bar{F}}_q(\mathbf{u}_q), \mathbf{M}_s(\mathbf{u}_s)\mathbf{\bar{F}}_s(\mathbf{u}_s)),
\end{equation}
where $\mathbf{u}_s=(x_s,y_s)$ and $\mathbf{u}_q=(x_q,y_q)$ denote spatial coordinates. The background similarity map $\mathbf{M}_{bg}$ is obtained in the same manner, and the query foreground confidence map is further computed as $\mathbf{M}_{conf} = \mathbf{M}_{fg}-\mathbf{M}_{bg}$.

In the auxiliary branch, we obtain the foreground distribution $\mathcal{N}(\boldsymbol{\mu}_{fg}, \boldsymbol{\sigma}^2_{fg})$ from the semantic bank and sample $P$ pseudo foreground prototypes $\{\mathbf{p}_{fg}^i\}_{i=1}^P$. We then compute the cosine similarity between $\mathbf{\bar{F}}_q$ and $\{\mathbf{p}_{fg}^i\}_{i=1}^P$. By maximizing and averaging along the prototype dimension, we obtain the maximum foreground similarity map $\mathbf{M}_{fg}^{max}$ and the average foreground similarity map $\mathbf{M}_{fg}^{avg}$:
\begin{equation}
\mathbf{M}_{fg}^{max}=\max_{i} \cos(\mathbf{\bar{F}}_q, \mathbf{p}_{fg}^i), \quad\mathbf{M}_{fg}^{avg}=\frac{1}{P} {\sum}_{i}^{P} \cos(\mathbf{\bar{F}}_q, \mathbf{p}_{fg}^i).
\end{equation}
The background similarity maps $\mathbf{M}_{bg}^{max}$ and $\mathbf{M}_{bg}^{avg}$ are computed in the same way, from which we further derive the max foreground confidence map $\mathbf{M}_{conf}^{max}$ and average foreground confidence map $\mathbf{M}_{conf}^{avg}$.

We combine the results from both branches to obtain the final foreground confidence map $\mathbf{\tilde{M}}_{conf} = \mathbf{M}_{conf}+\mathbf{M}_{conf}^{max}+\mathbf{M}_{conf}^{avg}$. Finally, we binarize $\mathbf{\tilde{M}}_{conf}$ using the adaptive threshold from \cite{sun2026bridging} to produce the query prediction $\mathbf{\tilde{M}}_{q}$.

\subsection{Loss Function}
We train our model on the main branch using the Binary Cross Entropy (BCE) loss:
\begin{equation}
\mathcal{L}_{main}=\text{BCE}(\mathbf{M}_{fg},\mathbf{M}_{bg},\mathbf{M}_q).
\end{equation}
Following \cite{sun2026bridging}, we additionally adopt the same self-segmentation loss $\mathcal{L}_{ssp}$ as in SSP \cite{fan2022self} to better supervise training. 

To prevent the spatial slots $\mathbf{\hat{S}}^i_{sp}$ and channel slots $\mathbf{\hat{S}}^i_{ch}$ from collapsing into similar representations, we normalize all slots and impose an orthogonality constraint at each stage:
\begin{equation}
\mathcal{L}_{orth}=\frac{1}{L_{s}}{\sum}_{i=1}^{L_{s}}\|\mathbf{\hat{S}}^{i}_{sp}(\mathbf{\hat{S}}^{i}_{sp})^T-\mathbf{I}\|_F^2 + \frac{1}{L_{c}}{\sum}_{i=1}^{L_{c}}\|\mathbf{\hat{S}}^{i}_{ch}(\mathbf{\hat{S}}^{i}_{ch})^T-\mathbf{I}\|_F^2.
\end{equation}

Finally, the total loss function is $\mathcal{L}=\mathcal{L}_{main}+\mathcal{L}_{ssp}+\lambda \mathcal{L}_{orth}$, where $\lambda=0.1$ is the weight parameter, and $\mathcal{L}_{ssp}$ is the same as that proposed in \cite{fan2022self}.

\begin{table*}[t]
    \caption{Mean-IoU of 1-shot and 5-shot results compared with previous FSS and CD-FSS methods on the CD-FSS benchmark. The best and second-best methods are highlighted in \textbf{bold} and \underline{underlined}, respectively.}
    \label{tab:performance}
    \centering
    \setlength{\tabcolsep}{5pt}
    \resizebox{\textwidth}{!}{
    \begin{tabular}{l|c|c|cc|cc|cc|cc|cc}
        \toprule
        \rule{0pt}{2.5ex}
        \multirow{2}*{Methods} & \multirow{2}*{Publication} & \multirow{2}*{Backbone} & \multicolumn{2}{c|}{Deepglobe} & \multicolumn{2}{c|}{ISIC} & \multicolumn{2}{c|}{Chest X-ray} & \multicolumn{2}{c|}{FSS-1000} & \multicolumn{2}{c}{Average} \\
        \cline{4-13}
        \rule{0pt}{2.5ex}
        & & & \multicolumn{1}{c}{1-shot} & 5-shot & \multicolumn{1}{c}{1-shot} & 5-shot & \multicolumn{1}{c}{1-shot} & 5-shot & \multicolumn{1}{c}{1-shot} & 5-shot & \multicolumn{1}{c}{1-shot} & 5-shot \\
        \midrule
        \multicolumn{13}{c}{Few-shot Semantic Segmentation Methods} \\
        \midrule
        RePRI \cite{boudiaf2021few} & CVPR-21 & Res-50 & 25.03 & 27.41 & 23.27 & 26.23 & 65.08 & 65.48 & 70.96 & 74.23 & 46.09 & 48.34 \\
        HSNet \cite{min2021hypercorrelation} & ICCV-21 & Res-50 & 29.65 & 35.08 & 31.20 & 35.10 & 51.88 & 54.36 & 77.53 & 80.99 & 47.57 & 51.38 \\
        SSP \cite{fan2022self} & ECCV-22 & Res-50 & 40.00 & 48.68 & 35.49 & 45.86 & 74.44 & 74.26 & 78.91 & 80.59 & 57.21 & 62.35 \\
        FPTrans \cite{zhang2022feature} & NIPS-22 & ViT-base & 38.36 & 49.30 & 48.65 & 60.37 & 80.92 & 82.91 & 80.74 & 83.65 & 62.17 & 69.06 \\
        HDMNet \cite{peng2023hierarchical} & CVPR-23 & Res-50 & 25.40 & 39.10 & 33.00 & 35.00 & 30.60 & 31.30 & 75.10 & 78.60 & 41.00 & 46.00 \\
        PerSAM \cite{zhang2023personalize}  & ICLR-24 & ViT-base & 36.08 & 40.65 & 23.27 & 25.33 & 29.95 & 30.05 & 60.92 & 66.53 & 37.56 & 40.64 \\
        VRP-SAM \cite{sun2024vrp} & CVPR-24 & ViT-base & 40.43 & 44.75 & 28.21 & 31.96 & 30.54 & 29.99 & 80.78 & 83.18 & 44.99 & 47.47 \\
        \midrule
        \multicolumn{13}{c}{Cross-domain Few-shot Semantic Segmentation Methods} \\
        \midrule
        PATNet \cite{lei2022cross} & ECCV-22 & Res-50 & 37.89 & 42.97 & 41.16 & 53.58 & 66.61 & 70.20 & 78.59 & 81.23 & 56.06 & 61.99 \\
        ABCDFSS \cite{herzog2024adapt} & CVPR-24 & Res-50 & 42.60 & 49.00 & 45.70 & 53.30 & 79.80 & 81.40 & 74.60 & 76.20 & 60.67 & 64.97 \\
        DRA \cite{su2024domain} & CVPR-24 & Res-50 & 41.29 & 50.12 & 40.77 & 48.87 & 82.35 & 82.31 & 79.05 & 80.40 & 60.86 & 65.42 \\
        APSeg \cite{he2024apseg} & CVPR-24 & ViT-base & 35.94 & 39.98 & 45.43 & 53.98 & 84.10 & 84.50 & 79.71 & 81.90 & 61.30 & 65.09 \\
        DMTNet \cite{chen2024cross} & IJCAI-24 & Res-50 & 40.14 & 51.17 & 43.55 & 52.30 & 73.74 & 77.30 & 81.52 & 83.28 & 59.74 & 66.01 \\
        APM \cite{tong2024lightweight} & NIPS-24 & Res-50 & 40.86 & 44.92 & 41.71 & 51.16 & 78.25 & 82.81 & 79.29 & 81.83 & 60.03 & 65.18 \\
        LoEC \cite{liu2025devil} & CVPR-25 & ViT-base & 42.12 & 51.48 & 52.91 & 62.43 & 83.94 & 84.12 & 81.05 & 83.69 & 65.01 & 70.43 \\
        SDRC \cite{tong2025self} & ICML-25 & ViT-base & 43.15 & 46.83 & 46.57 & 55.02 & 82.86 & 84.79 & 80.31 & 82.55 & 63.22 & 67.30 \\
        DFN \cite{tong2025adapter} & ICML-25 & ViT-base & 39.45 & 47.67 & 50.36 & 58.53 & 83.18 & \textbf{87.14} & \underline{82.97} & 85.72 & 63.99 & 69.77 \\
        ISA \cite{fan2025adapting}  & ICCV-25 & Res-50 & 44.30 & 52.70 & 37.20 & 56.10 & 83.40 & 86.30 & 78.80 & \underline{86.00} & 60.90 & 70.30 \\
        HSL \cite{sun2026bridging} & AAAI-26 & ViT-base & \underline{45.77} & \underline{54.56} & \underline{59.36} & \underline{64.62} & \underline{85.95} & 86.25 & 81.89 & 83.84 & \underline{68.24} & \underline{72.32} \\
        \midrule
        \textbf{DHANet(Ours)} & Ours & ViT-base & \textbf{49.79} & \textbf{56.13} & \textbf{64.25} & \textbf{67.38} & \textbf{86.06} & \underline{86.87} & \textbf{83.61} & \textbf{86.67} & \textbf{70.93} & \textbf{74.26} \\
        \bottomrule
    \end{tabular}}
\end{table*}

\section{Experiments}
\label{sec:exper}

\subsection{Experiment Setup}
\textbf{Datasets and Metric.}
Following the experimental setup of PATNet \cite{lei2022cross}, we use the PASCAL VOC 2012 \cite{everingham2010pascal} dataset augmented with SBD \cite{hariharan2011semantic} as the source domain for training. Subsequently, the trained model is evaluated on four target datasets: Deepglobe \cite{demir2018deepglobe}, ISIC \cite{codella2019skin, tschandl2018ham10000}, Chest X-ray \cite{candemir2013lung, jaeger2013automatic}, and FSS-1000 \cite{li2020fss}. Deepglobe \cite{demir2018deepglobe} is a satellite remote sensing dataset comprising dense annotations for 7 categories: areas of urban, agriculture, rangeland, forest, water, barren, and unknown. ISIC \cite{codella2019skin, tschandl2018ham10000} is a dermoscopic skin lesion dataset covering 3 major types of skin lesions. Chest X-ray \cite{candemir2013lung, jaeger2013automatic} is a tuberculosis diagnosis X-ray dataset, where the grayscale images exhibit substantial style gaps compared to the natural images. FSS-1000 \cite{li2020fss} contains 1,000 daily object categories with 10 annotated images per class.

We adopt mean Intersection over Union (mIoU) as the evaluation metric and report the average results over 5 runs with diﬀerent random seeds.

\noindent
\textbf{Implementation Details.}
Consistent with previous works \cite{liu2025devil, sun2026bridging}, we adopt ViT-B/16 \cite{dosovitskiy2020image} as our backbone and resize all images to $480 \times 480$. During training, we follow FPTrans \cite{zhang2022feature} to augment all images. We use SGD to optimize our model for 5 epochs, with a momentum of 0.9, a weight decay of 5e-4, a batch size of 6, and a constant learning rate of 1e-3. Following most works, we also perform fine-tuning using the support set from the first episode of each class in the target domain. During fine-tuning, we sample 12 episodes from the support set augmented using the strategy in \cite{herzog2024adapt} and train for one epoch. We set the number of aggregation stages $L_s$ in HSA to 2, with corresponding slot numbers of $(10, 20)$. Similarly, the number of stages $L_c$ in HCA is set to 2 with $(32, 64)$ slots. The number of sampled pseudo-prototypes $P$ in OPSB is set to 20. All experiments are conducted on a single NVIDIA GeForce RTX 4090 GPU.

\begin{table}[t]
    \begin{minipage}{0.625\linewidth}
        \caption{Ablation studies for each component of our method.}
        \vspace{-1ex}
        \label{tab:component}
        \centering
        \setlength{\tabcolsep}{1.8mm}
        \resizebox{\linewidth}{!}{
        \begin{tabular}{ccc|cccc|c}
            \toprule
            HSA & HCA & OPSB & Deepglobe & ISIC & Chest & FSS-1000 & Average \\
            \midrule
            $\times$ & $\times$ & $\times$ & 42.52 & 55.75 & 84.60 & 83.23 & 66.53 \\
            \checkmark & $\times$ & $\times$ & 45.24 & 59.23 & 86.07 & 83.16 & 68.43 \\
            $\times$ & \checkmark & $\times$ & 44.48 & 58.19 & 85.77 & 83.17 & 67.90 \\
            \checkmark & \checkmark & $\times$ & 46.68 & 60.43 & 86.52 & 83.25 & 69.22 \\
            \checkmark & \checkmark & \checkmark & 49.79 & 64.25 & 86.06 & 83.61 & \textbf{70.93} \\ 
            \bottomrule
        \end{tabular}}
    \end{minipage}
    \hfill
    \begin{minipage}{0.325\linewidth}
        \caption{Effects of components in HSA and HCA.}
        \vspace{-1ex}
        \label{tab:cluster_component}
        \centering
        \setlength{\tabcolsep}{3.2mm}
        \resizebox{\linewidth}{!}{
        \begin{tabular}{c|c}
            \toprule
             & mean-IoU \\
            \midrule
            All & \textbf{69.22} \\
            w/o SlotAttn & 67.39 \\
            w/o CrossAttn & 67.81 \\
            w/o $\mathcal{L}_{orth}$ & 68.33 \\
            \bottomrule
        \end{tabular}}
    \end{minipage}
    \vspace{-2ex}
\end{table}

\subsection{Comparison with State-of-the-art Methods}
\begin{figure}[tb]
    \centering
    \includegraphics[width=0.97\linewidth]{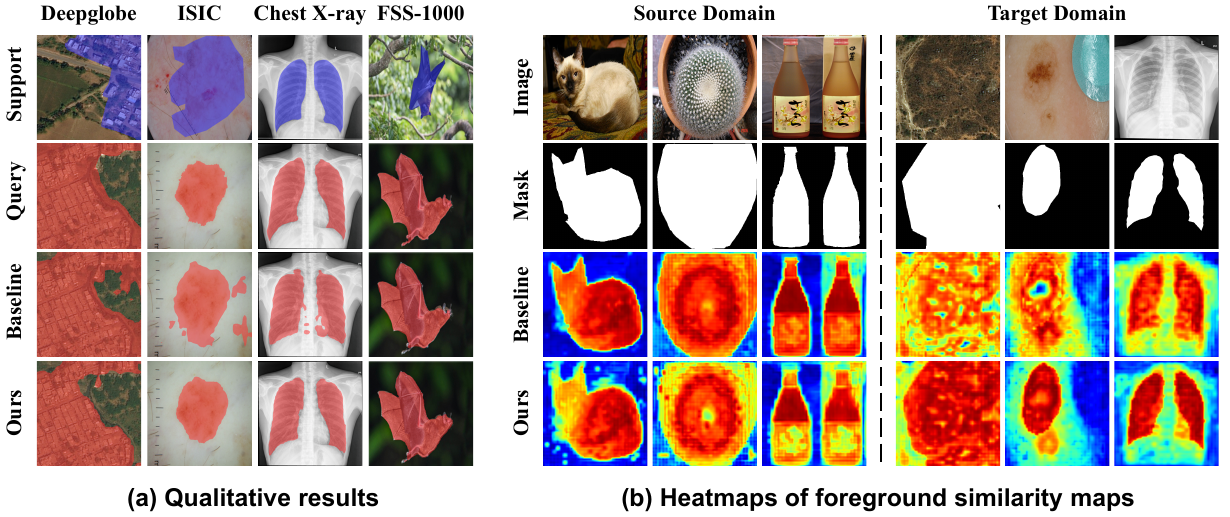}
    \vspace{-1ex}
    \caption{(a) Qualitative results of our model for 1-way 1-shot setting. (b) The heatmaps of the foreground similarity maps show that our method can extract hierarchical semantic features to alleviate semantic over-alignment.}
    \vspace{-1ex}
    \label{fig:results}
\end{figure}

In \cref{tab:performance}, we compare our method with existing FSS and CD-FSS methods, achieving remarkable performance advantages under both 1-shot and 5-shot settings. Specifically, our method surpasses the state-of-the-art method HSL \cite{sun2026bridging} by 2.69$\%$ and 1.94$\%$ under the 1-shot and 5-shot settings, respectively, and our 1-shot performance even exceeds the 5-shot performance of all other methods except HSL. Moreover, our method shows more pronounced advantages on target domains with larger category differences from the source domain. For instance, the 1-shot performance on Deepglobe and ISIC surpasses HSL by 4.02$\%$ and 4.89$\%$, respectively. These results clearly demonstrate the effectiveness of our method. Additionally, we conduct a fair comparison with some CD-FSS methods under the setup of IFA \cite{nie2024cross}. Please refer to our supplementary material for more details. We also present some qualitative results of our method for 1-way 1-shot setting in \cref{fig:results}(a).

\subsection{Ablation Studies}
\textbf{Effects of Each Component.}
As shown in \cref{tab:component}, we analyze each component under the 1-shot setting to validate the effectiveness of each design. The baseline builds on SSP \cite{fan2022self} by applying global style randomization from \cite{liu2025devil} and target-domain fine-tuning to alleviate style gaps, and adopts the adaptive thresholding strategy in HSL for segmentation. HSA mitigates semantic over-alignment by constructing hierarchical semantic features, improving the baseline's average performance by 1.90$\%$. HCA alleviates attribute over-alignment caused by hard channel-wise alignment via hierarchical channel aggregation, yielding a 1.37$\%$ average gain over the baseline. HSA and HCA are complementary, jointly bringing a 2.69$\%$ mIoU gain. In addition, OPSB extracts high-confidence features from the query prediction and builds a class semantic bank to provide extra support information, further improving performance by 1.71$\%$. These results fully demonstrate the effectiveness of each component.

\noindent
\textbf{Effects of Components in HSA and HCA.}
We conduct ablation studies to investigate the importance of each component in the aggregation module, as shown in \cref{tab:cluster_component}. Slot attention partitions features into distinct groups and aggregates information from them. Without it, there is no guarantee that individual slots aggregate distinct features, leading to performance degradation. Cross attention injects sample priors into the slots, enabling the aggregated slots to establish support–query correspondences. Removing cross attention causes unstable channel alignment in HCA, thus compromising performance. The orthogonal loss is applied to encourage the model in capturing differences among image features corresponding to different slots. Discarding this loss results in limited semantic discriminability for the features and a subsequent drop in performance. The results demonstrate the importance of these components.

\begin{table}[t]
    \begin{minipage}{0.48\linewidth}
        \caption{Effect of the number of aggregation stages in HSA.}
        \vspace{-1ex}
        \label{tab:stage_hsa}
        \centering
        \setlength{\tabcolsep}{3.2mm}
        \resizebox{\linewidth}{!}{
            \begin{tabular}{c|cccc|c}
                \toprule
                $L_s$ & 10 & 20 & 40 & 80 & mean-IoU \\
                \midrule
                1 & \checkmark & $\times$ & $\times$ & $\times$ & 67.92 \\
                2 & \checkmark & \checkmark & $\times$ & $\times$ & 68.43 \\
                3 & \checkmark & \checkmark & \checkmark & $\times$ & \textbf{68.44} \\
                4 & \checkmark & \checkmark & \checkmark & \checkmark & 68.11 \\
                \bottomrule
            \end{tabular}}
    \end{minipage}
    \hfill
    \begin{minipage}{0.48\linewidth}
        \caption{Effect of the number of aggregation stages in HCA.}
        \vspace{-1ex}
        \label{tab:stage_hca}
        \centering
        \setlength{\tabcolsep}{3.2mm}
        \resizebox{\linewidth}{!}{
            \begin{tabular}{c|cccc|c}
                \toprule
                $L_c$ & 32 & 64 & 128 & 256 & mean-IoU \\
                \midrule
                1 & \checkmark & $\times$ & $\times$ & $\times$ & 68.84 \\
                2 & \checkmark & \checkmark & $\times$ & $\times$ & \textbf{69.22} \\
                3 & \checkmark & \checkmark & \checkmark & $\times$ & 69.04 \\
                4 & \checkmark & \checkmark & \checkmark & \checkmark & 68.93 \\
                \bottomrule
            \end{tabular}}
    \end{minipage}
    \vspace{-2ex}
\end{table}

\noindent
\textbf{Number of Aggregation Stages.}
We conduct ablation studies under the 1-shot setting to validate the necessity of multi-level aggregation. Specifically, we gradually increase the number of aggregation stages in HSA and HCA, respectively, with the number of slots increasing progressively at each stage. The results are shown in \cref{tab:stage_hsa} and \cref{tab:stage_hca}. The best performance in HSA is achieved with 3 stages of spatial aggregation. However, since 2 stages yield almost identical performance with lower computational overhead, we set the number of aggregation stages $L_s$ to 2. Similarly, since 2 stages of channel aggregation in HSA achieve the best performance, we also set the number of aggregation stages $L_c$ to 2. These results demonstrate that hierarchical aggregation is necessary. However, performance does not continue to improve as the number of aggregation stages further increases. A possible reason is that excessively fine-grained aggregation and orthogonal constraints may instead disrupt the original category hierarchy, leading to negative effects.

\noindent
\textbf{Query Prediction Utilization Strategies.}
To validate the effectiveness of our probabilistic semantic bank, we compare different strategies for utilizing query information during testing, as reported in \cref{tab:opsb}. Existing CD-FSS methods \cite{lei2022cross, chen2024cross} typically leverage predicted pseudo query masks to further fine-tune the model during inference. However, such online fine-tuning incurs additional computational overhead and is prone to knowledge forgetting. In contrast, directly computing pseudo prototypes from pseudo query masks and maintaining as well as updating this deterministic prototype online can provide stable additional support information. Our method instead constructs a probabilistic prototype and samples multiple pseudo prototypes, thereby offering more diverse support information and achieving significant performance gains.

\begin{figure}[t]
\centering
    \includegraphics[width=0.97\linewidth]{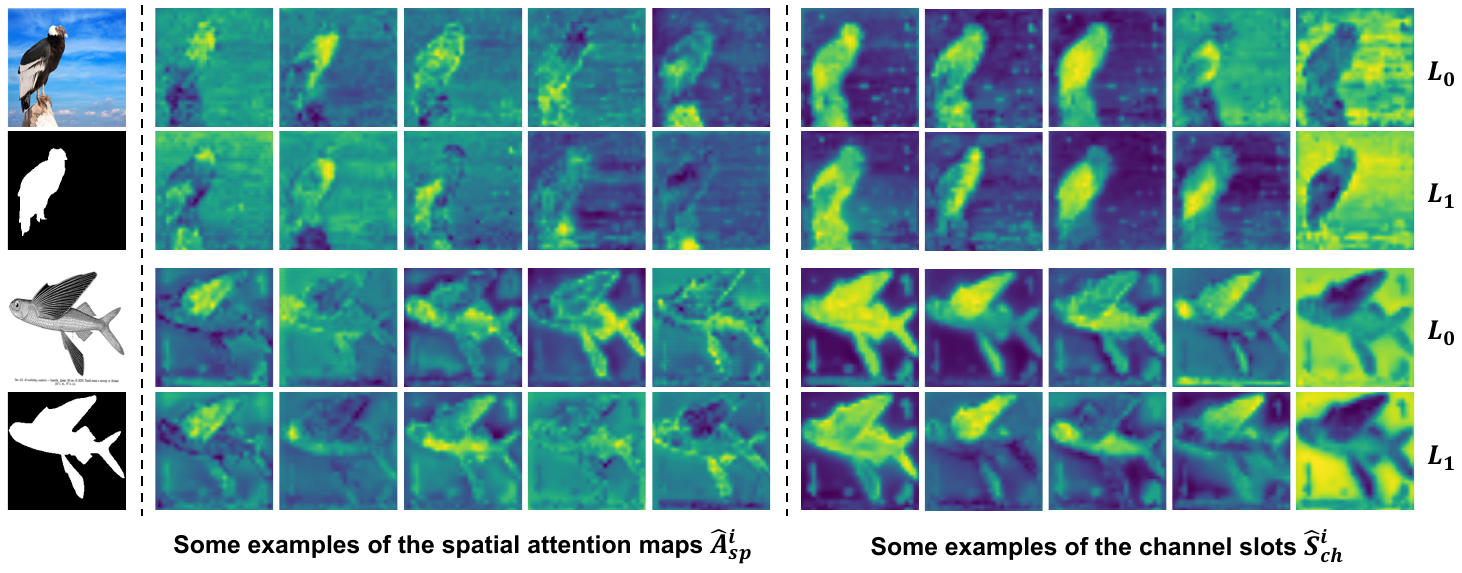}
    \vspace{-1ex}
    \caption{Visualization of spatial and channel slots demonstrates that multi-stage slots facilitate constructing features with hierarchical semantic and attribute enhancement.}
    \vspace{-1ex}
    \label{fig:slot_vis}
\end{figure}

\begin{table}[t]
    \begin{minipage}{0.6\linewidth}
        \caption{Comparison of different query prediction utilization strategies.}
        \vspace{-1ex}
        \label{tab:opsb}
        \centering
        \setlength{\tabcolsep}{1.2mm}
        \resizebox{\linewidth}{!}{    
            \begin{tabular}{c|cccc|c}
                \toprule
                & Deepglobe & ISIC & Chest & FSS & Average \\
                \midrule
                Fine-tuning & 44.75 & 57.29 & 86.41 & 83.07 & 67.88 \\
                Deterministic prototype & 47.87 & 62.89 & 85.78 & 83.26 & 69.95 \\
                Probabilistic prototype & 49.79 & 64.25 & 86.06 & 83.61 & \textbf{70.93} \\
                \bottomrule
            \end{tabular}}
    \end{minipage}
    \hfill
    \begin{minipage}{0.36\linewidth}
        \caption{Model efficiency and performance under 1-shot setting.}
        \vspace{-1ex}
        \label{tab:efficiency}
        \centering   
        \setlength{\tabcolsep}{1.2mm}
        \resizebox{\linewidth}{!}{ 
            \begin{tabular}{c|ccc}
                \toprule
                Method & FLOPs (G) & FPS & mean-IoU\\
                \midrule
                HSL \cite{sun2026bridging} & 186.52 & 27.25 & 68.24 \\
                Ours & 176.48 & 30.77 & 70.93 \\
                \bottomrule
            \end{tabular}
        }
    \end{minipage}
    \vspace{-1ex}
\end{table}

\subsection{Analysis and Visualizations}
\textbf{Model Efficiency.}
We compare model efficiency with the current state-of-the-art method HSL \cite{sun2026bridging} under the 1-shot setting, including computational complexity (FLOPs) and inference speed (FPS). As shown in \cref{tab:efficiency}, our method demonstrates both better model efficiency and segmentation performance.

\noindent
\textbf{Visualization of Slots.}
To validate the construction process of our hierarchical features, we visualize the spatial and channel slots at each stage. Specifically, we visualize the attention maps of spatial slots and the spatial activation maps of channel slots after restoring their spatial structure, as shown in \cref{fig:slot_vis}. Here, $L_0$ denotes the coarse-grained stage (with fewer slots), and $L_1$ denotes the fine-grained stage. Different spatial slots attend to distinct semantic regions, with those in the fine-grained stage capturing more detailed semantics. Furthermore, channel slots activate regions with similar attributes. Since the coarse-grained channel slots aggregate more channels, their activated regions are typically larger than those of the fine-grained slots. These visualizations intuitively demonstrate that multi-stage slots facilitate the construction of features with hierarchical semantic and attribute enhancement.

\noindent
\textbf{Mitigating semantic over-alignment.}
Semantic over-alignment makes the learned features excel at distinguishing categories only at the source-domain granularity, while lacking the ability to discriminate target-domain semantics. We utilize ground truth masks to extract foreground prototypes and calculate similarity maps, with the corresponding heatmaps visualized in \cref{fig:results}(b). The baseline overfits to extracting highly consistent features for source-domain categories, failing to effectively distinguish foreground from background in the target domain. In contrast, our method focuses on extracting more hierarchical features, enabling generalization to diverse domains.

\begin{wraptable}{r}{0.4\textwidth}
    \vspace{-4ex}
    \caption{Effects of our mehod on channel correlation.}
    \vspace{2ex}
    \label{tab:pearson}
    \centering
    \setlength{\tabcolsep}{1.2mm}
    \resizebox{\linewidth}{!}{    
        \begin{tabular}{c|cccc}
            \toprule
            & Deepglobe & ISIC & Chest & FSS \\
            \midrule
            Baseline & 0.2713 & 0.3149 & 0.3307 & 0.2817  \\
            Ours & 0.2032 & 0.2366 & 0.2257 & 0.2318 \\
            \bottomrule
        \end{tabular}}
    \vspace{-4ex}
\end{wraptable}

\noindent
\textbf{Mitigating attribute over-alignment.}
Attribute over-alignment causes the degradation of source-insensitive attributes, where all channels tend to represent source-sensitive attributes, leading to attribute redundancy, \ie, high inter-channel correlation. Following \cite{wang2019cop}, we use the Pearson correlation coefficient to measure the degree of inter-channel redundancy. As shown in \cref{tab:pearson}, our hierarchical channel aggregation mitigates channel misalignment and reduces inter-channel correlation, enabling features to capture more diverse attributes for adaptation to different domains.

\section{Conclusion}
\label{sec:conclusion}
In this paper, we propose a Dual Hierarchical Aggregation Network (DHANet) to simultaneously alleviate semantic over-alignment and attribute over-alignment in CD-FSS. Specifically, we propose a Hierarchical Spatial Aggregation (HSA) module, which aligns features at multiple granularities through hierarchical aggregation along the spatial dimension, thereby adapting to target domains with varying segmentation granularities. Furthermore, we introduce a Hierarchical Channel Aggregation (HCA) module that mitigates the rigid channel-wise hard alignment via hierarchical aggregation along the channel dimension, thus preserving sensitivity to diverse attributes. HSA and HCA effectively alleviate semantic and attribute over-alignment, respectively. Finally, we present an Online Probabilistic Semantic Bank (OPSB) module to address the scarcity of support information during inference. Extensive experiments demonstrate that our DHANet achieves state-of-the-art performance across multiple CD-FSS datasets.

\section*{Acknowledgments}
This work was supported in part by the National Natural Science Foundation of China under the Grants No. 62371235 and No. U25A20444, in part by the Key Research and Development Plan of Jiangsu Province under Grant No. BE2023008-2.

%
%
\bibliographystyle{splncs04}
\bibliography{main}
\end{document}


\title{Hierarchical Spatial and Channel Aggregation for Cross-domain Few-shot Segmentation} 

\titlerunning{Hierarchical Spatial and Channel Aggregation for CD-FSS}

\author{Sujun Sun\inst{1,2}\orcidlink{0009-0002-0647-1148} \and
Mingwu Ren\inst{1,2}\orcidlink{0000-0001-5576-3281} \and
Haofeng Zhang\inst{1,2}\textsuperscript{(\Letter)}\orcidlink{0000-0002-4039-7618}}

\authorrunning{S. Sun et al.}

\institute{School of Computer Science and Engineering, Nanjing University of Science and Technology, China \and
State Key Laboratory of Intelligent Manufacturing of Advanced Construction Machinery, China\\
\email{\{egg, renmingwu, zhanghf\}@njust.edu.cn}}

\maketitlesupplementary

\section{Slot Attention}
\label{sec:slot_attn}
Slot attention \cite{locatello2020object} is an iterative attention module that adaptively aggregates input features into a finite number of slots, and it is widely used in unsupervised/weakly supervised object-centric representation learning and grouping tasks \cite{ma2023attrseg, liao2025forla}. Its core idea is to competitively explain the given $N_f$ input features $\mathbf{F}\in \mathbb{R}^{N_f\times C}$ using $N_s$ learnable slots $\mathbf{S}\in \mathbb{R}^{N_s\times C}$, thereby forming an aggregated representation within each slot that focuses on a specific object or part.

First, the input features are normalized and linearly projected to generate keys and values:
\begin{equation}
    \mathbf{K}=\operatorname{LN}(\mathbf{F})\mathbf{W}_K\in\mathbb{R}^{N_f\times C},\quad\mathbf{V}=\operatorname{LN}(\mathbf{F})\mathbf{W}_V\in\mathbb{R}^{N_f\times C},
\end{equation}
where $\operatorname{LN}$ denotes Layer Normalization, and $\mathbf{W} \in \mathbb{R}^{C \times C}$ is a learnable linear projection matrix.

\noindent\textbf{Competitive attention allocation and aggregation.}
Slot attention updates $\mathbf{S}$ iteratively over $T$ iterations. At the $t$-th iteration, the slots are first projected into queries:
\begin{equation}
    \mathbf{Q}^{(t)}=\operatorname{LN}(\mathbf{S}^{(t)})\mathbf{W}_Q\in\mathbb{R}^{N_s\times C}.
\end{equation}
Subsequently, the attention logits between the slots and the input features are computed as:
\begin{equation}
    \mathbf{L}^{(t)}=\frac{\mathbf{Q}^{(t)}\mathbf{K}^\top}{\sqrt{C}}\in\mathbb{R}^{N_s\times N_f}.
\end{equation}
To enable competitive allocation of each feature across different slots, softmax normalization along the slots dimension is applied to the attention logits, yielding the attention map $\mathbf{A}^{(t)}$. This normalization ensures that the slots compete to explain the input features, encouraging different slots to bind to distinct parts of the input.

Next, the inputs for the slot update step are obtained via weighted aggregation:
\begin{equation}
    \mathbf{U}^{(t)}=\mathbf{A}^{(t)}\mathbf{V}\in\mathbb{R}^{N_{s}\times C}.
\end{equation}

\noindent\textbf{Slot update.}
Slot attention employs a shared-parameter gated recurrent unit (GRU) to write the aggregated information back into the slots:
\begin{equation}
    \mathbf{S}^{(t)}\leftarrow \operatorname{GRU}(\mathbf{S}^{(t)},\mathbf{U}^{(t)}).
\end{equation}
Then, a feed-forward network (FFN) is applied for residual refinement:
\begin{equation}
    \mathbf{S}^{(t+1)}=\mathbf{S}^{(t)}+\operatorname{FFN}\left(\operatorname{LN}(\mathbf{S}^{(t)})\right).
\end{equation}

After $T$ iterations, the final slot representations $\mathbf{S}^{(T)}$ are obtained, which can serve as object- or part-level representations for downstream tasks.

\begin{table}[t]
    \caption{Mean-IoU of 1-shot and 5-shot results compared with previous CD-FSS methods under the setup of IFA \cite{nie2024cross}. The best and second-best methods are highlighted in \textbf{bold} and \underline{underlined}, respectively.}
    \label{tab:performance_ifa}
    \centering
    \setlength{\tabcolsep}{3pt}
    \resizebox{\textwidth}{!}{
    \begin{tabular}{l|c|cc|cc|cc|cc|cc}
        \toprule
        \rule{0pt}{2.5ex}
        \multirow{2}*{Methods} & \multirow{2}*{Publication} & \multicolumn{2}{c|}{Deepglobe} & \multicolumn{2}{c|}{ISIC} & \multicolumn{2}{c|}{Chest X-ray} & \multicolumn{2}{c|}{FSS-1000} & \multicolumn{2}{c}{Average} \\
        \cline{3-12}
        \rule{0pt}{2.5ex}
        & & \multicolumn{1}{c}{1-shot} & 5-shot & \multicolumn{1}{c}{1-shot} & 5-shot & \multicolumn{1}{c}{1-shot} & 5-shot & \multicolumn{1}{c}{1-shot} & 5-shot & \multicolumn{1}{c}{1-shot} & 5-shot \\
        \midrule
        IFA \cite{nie2024cross} & CVPR-24 & 50.6 & 58.8 & 66.3 & 69.8 & 74.0 & 74.6 & 80.1 & 82.4 & 67.8 & 71.4 \\
        GPRN \cite{peng2025sam} & AAAI-25 & 51.7 & 59.3 & 66.8 & 72.2 & 87.0 & 87.1 & 81.1 & 82.6 & 71.7 & 75.3 \\
        DATO \cite{li2025dual} & CVPR-25 & 51.1 & 59.3 & 68.8 & 70.3 & 79.6 & 81.1 & 81.8 & 84.6 & 70.3 & 73.8 \\
        SDRC \cite{tong2025self} & ICML-25 & 51.1 & 59.4 & 69.7 & 72.5 & 84.1 & 87.2 & 83.1 & 85.7 & 72.0 & 76.2 \\
        DFN \cite{tong2025adapter} & ICML-25 & 51.3 & 59.2 & 68.5 & 71.4 & 86.1 & \textbf{91.6} & 82.6 & \underline{87.9} & 72.1 & 77.5 \\
        HSL \cite{sun2026bridging} & AAAI-26 & \underline{51.8} & \underline{59.5} & \underline{75.5} & \underline{78.4} & \underline{88.0} & 88.5 & \underline{85.5} & 86.2 & \underline{75.2} & \underline{78.1} \\
        \textbf{DHANet(Ours)} & - & \textbf{53.9} & \textbf{60.0} & \textbf{77.0} & \textbf{80.3} & \textbf{88.3} & \underline{89.1} & \textbf{88.4} & \textbf{90.4} & \textbf{76.9} & \textbf{80.0}\\
        \bottomrule
    \end{tabular}}
\end{table}

\section{Comparison with More Methods under Special Setting}
\label{sec:ifa}
IFA \cite{nie2024cross} calculates the mean-IoU by treating all samples in a batch as the same class, with the batch size set to 96. The results obtained in this manner fall between mean-IoU and FB-IoU, and tend to be higher than the true mean-IoU. Furthermore, IFA fine-tunes on the target domain support set for 20 epochs, but the samples drawn in each epoch are not guaranteed to be identical, \ie, it uses more data than a single 1-shot or 5-shot annotated set for fine-tuning. Some methods \cite{peng2025sam, li2025dual} strictly follow this setting, while others \cite{tong2025self, tong2025adapter, sun2026bridging} additionally report results under this setting. For a fair comparison, we also conduct experiments under the setting of IFA. As shown in \cref{tab:performance_ifa}, our method significantly outperforms existing methods under this setting.

\section{More Ablation Studies}
\label{sec:more_ablation}
We provide more ablation studies, including additional module designs and hyperparameter analysis experiments. All experiments are conducted under 1-shot setting unless stated otherwise.

\begin{table}[ht]
    \caption{Performance comparison on the ResNet-50 backbone.}
    \label{tab:performance_resnet}
    \centering
    \setlength{\tabcolsep}{2.4mm}
    \scalebox{.85}{
        \begin{tabular}{l|cccc|c}
            \toprule
            Methods & Deepglobe & ISIC & Chest X-ray & FSS-1000 & Average \\
            \midrule
            LoEC \cite{liu2025devil} & 44.10 & 38.21 & 81.02 & 78.51 & 60.46 \\
            DFN \cite{tong2025adapter} & 45.66 & 36.30 & 85.21 & 80.73 & 61.98 \\
            HSL \cite{sun2026bridging} & 46.13 & 48.01 & 84.57 & 78.22 & \underline{64.23} \\
            \midrule
            Baseline & 43.38 & 43.99 & 84.20 & 77.58 & 62.29 \\
            \textbf{DHANet(Ours)} & 49.92 & 49.88 & 85.00 & 79.29 & \textbf{66.02}\\
            \bottomrule
        \end{tabular}}
\end{table}

\noindent
\textbf{Generalization to the ResNet-50 backbone.}
We further adopt ResNet-50 \cite{he2016deep} as the backbone to validate the effectiveness of our method across different architectures. As shown in \cref{tab:performance_resnet}, our method also significantly improves performance over the baseline, outperforming existing state-of-the-art methods.

\begin{table}[ht]
    \caption{Comparison of different attribute prototype fusion strategies.}
    \label{tab:attr_fusion}
    \centering
    \setlength{\tabcolsep}{2.4mm}
    \scalebox{.85}{
        \begin{tabular}{l|cccc|c}
            \toprule
            Strategy & Deepglobe & ISIC & Chest X-ray & FSS-1000 & Average \\
            \midrule
            w/o HCA & 45.24 & 59.23 & 86.07 & 83.16 & 68.43 \\
            RMAP & 45.28 & 59.40 & 85.67 & 83.01 & 68.34 \\
            Concat & 46.68 & 60.43 & 86.52 & 83.25 & \textbf{69.22} \\
            \bottomrule
        \end{tabular}}
\end{table}

\noindent
\textbf{Attribute prototype fusion strategies.}
We compare different strategies for fusing attribute prototypes with original features in HCA, as reported in \cref{tab:attr_fusion}. In HSA, region filling via RMAP along the spatial dimension enhances regional consistency, which is beneficial for the segmentation task. However, in HCA, filling each attribute prototype into its corresponding channel position via RMAP along the channel dimension inadvertently increases inter-channel correlation and undermines attribute diversity, resulting in performance even worse than without hierarchical channel aggregation. In contrast, concatenating attribute prototypes with features along the channel dimension preserves both original and hierarchical attribute information, leading to improved performance.

\begin{table}[ht]
    \caption{Ablation studies for the effects of low-level feature $\mathbf{F}^l$.}
    \label{tab:low_level}
    \centering
    \setlength{\tabcolsep}{2.4mm}
    \scalebox{.85}{
        \begin{tabular}{l|cccc|c}
            \toprule
            Strategy & Deepglobe & ISIC & Chest X-ray & FSS-1000 & Average \\
            \midrule
            w/o $\mathbf{F}^l$ & 46.12 & 60.23 & 85.94 & 83.08 & 68.84 \\
            HSA + $\mathbf{F}^l$ & 46.33 & 60.15 & 86.48 & 83.16 & 69.03 \\
            HCA + $\mathbf{F}^l$ & 46.68 & 60.43 & 86.52 & 83.25 & \textbf{69.22} \\
            Both + $\mathbf{F}^l$ & 46.36 & 60.07 & 86.64 & 83.01 & 69.02 \\
            \bottomrule
        \end{tabular}}
\end{table}

\noindent
\textbf{Effects of low-level feature $\mathbf{F}^l$.}
Following \cite{liu2025devil, sun2026bridging}, we introduce the shallow low-level feature $\mathbf{F}^l$ to provide more detailed information and investigate the effects of incorporating $\mathbf{F}^l$ in different modules. As shown in \cref{tab:low_level}, introducing $\mathbf{F}^l$ into HSA and HCA individually further improves performance, whereas introducing $\mathbf{F}^l$ into both modules simultaneously leads to performance degradation. A possible reason is that excessive low-level features weaken the semantic information of high-level features.

\begin{figure}[t]
\centering
    \includegraphics[width=0.98\linewidth]{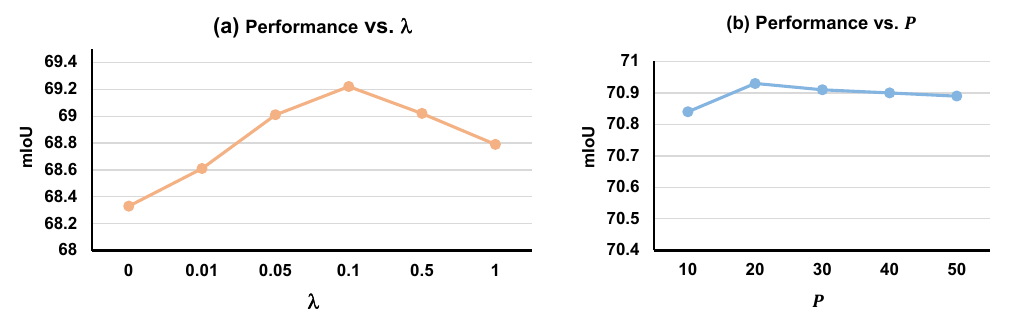}
    \caption{Parameter studies. (a) The mIoU curves over different loss weight $\lambda$. (b) The mIoU curves over different sampling numbers $P$ in OPSB.}
    \label{fig:param_line}
\end{figure}

\begin{table}[ht]
    \caption{Ablation studies for the effects of different foreground confidence maps.}
    \label{tab:conf_map}
    \centering
    \setlength{\tabcolsep}{2.4mm}
    \scalebox{.85}{
        \begin{tabular}{l|cccc|c}
            \toprule
             & Deepglobe & ISIC & Chest X-ray & FSS-1000 & Average \\
            \midrule
            w/o OPSB & 46.68 & 60.43 & 86.52 & 83.25 & 69.22 \\
            Only $\mathbf{M}_{conf}^{max}$ & 48.24 & 63.89 & 85.98 & 83.49 & 70.40 \\
            Only $\mathbf{M}_{conf}^{avg}$ & 47.75 & 62.84 & 85.99 & 83.46 & 70.01 \\
            $\mathbf{M}_{conf}^{max}$ + $\mathbf{M}_{conf}^{avg}$ & 49.79 & 64.25 & 86.06 & 83.61 & \textbf{70.93} \\
            \bottomrule
        \end{tabular}}
\end{table}

\noindent
\textbf{Effects of different foreground confidence maps.}
In \cref{tab:conf_map}, we compare the performance of using different foreground confidence maps in OPSB. Specifically, $\mathbf{M}_{conf}^{max}$ provides additional support from the pseudo-prototype that is most similar to the query, which is more accurate but also more susceptible to noise. Conversely, $\mathbf{M}_{conf}^{avg}$ approximates using a deterministic prototype, offering conservative yet stable additional support. Combining both maps allows them to complement each other, thereby achieving further performance improvements.

\noindent
\textbf{Effects of the loss weight $\lambda$.}
We set the loss weight $\lambda$ of $\mathcal{L}_{orth}$ to different values to explore their effects and show the results in \cref{fig:param_line}(a). Introducing $\mathcal{L}_{orth}$ helps prevent the homogenization of slots across stages, and the best performance is achieved when $\lambda$ is set to 0.1.

\noindent
\textbf{Effects of the sampling number $P$.}
As shown in \cref{fig:param_line}(b), we evaluate the model performance with different numbers of sampled pseudo-prototypes in OPSB. The results show that setting $P$ to 20 achieves the best performance. However, when $P$ is set to other values, the performance exhibits only marginal differences from the best, indicating that the model is not sensitive to $P$.

\begin{figure}[t]
\centering
    \includegraphics[width=1.0\linewidth]{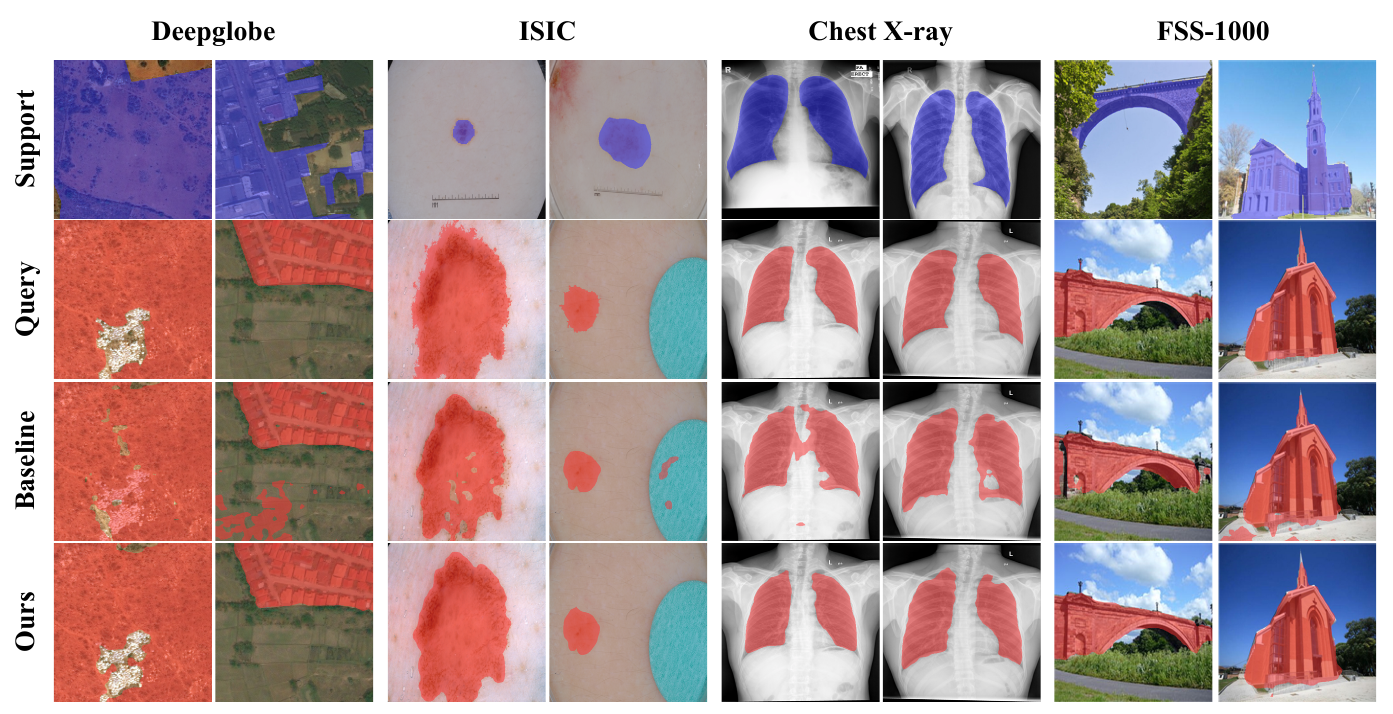}
    \caption{More qualitative results of our method for 1-way 1-shot segmentation. The support labels are highlighted in blue, while the predictions and ground truth of query images are presented in red.}
    \label{fig:more_res}
\end{figure}

\begin{figure}[t!]
\centering
    \includegraphics[width=1.0\linewidth]{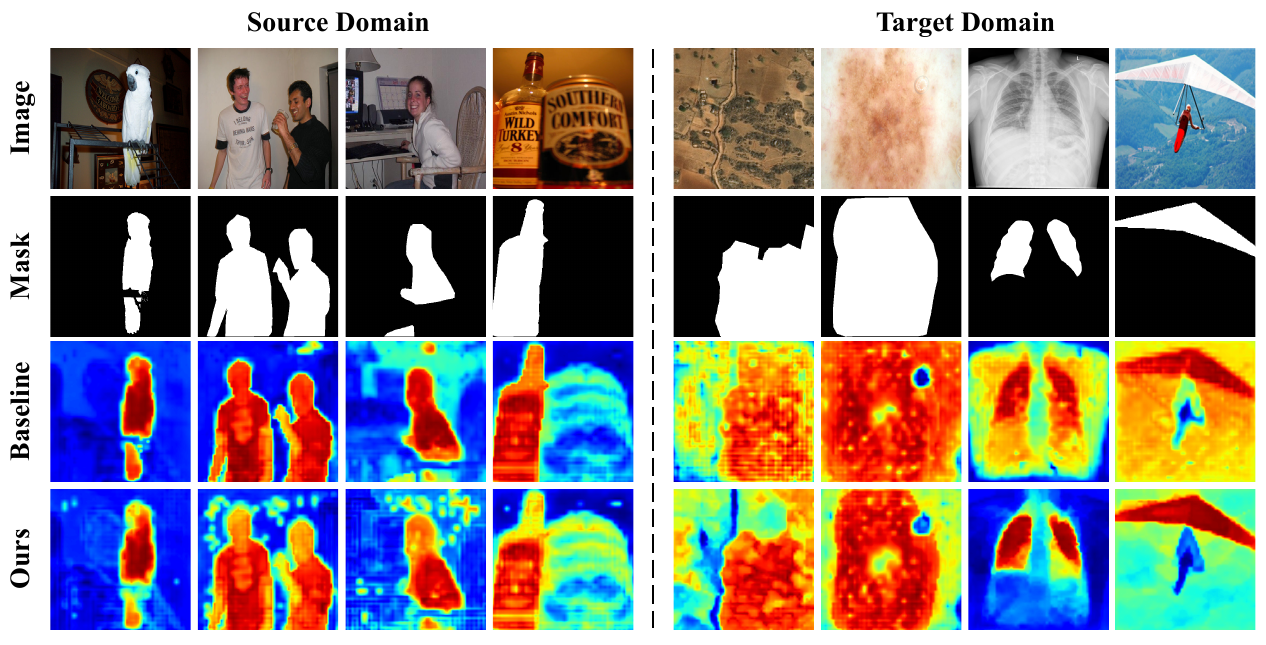}
    \caption{The heatmaps of the foreground similarity maps for source and target domains demonstrate that our method can extract hierarchical semantic features to alleviate semantic over-alignment.}
    \label{fig:more_simmap}
\end{figure}

\section{More Visualizations}
\label{sec:more_vis}

\noindent
\textbf{Qualitative results.}
In \cref{fig:more_res}, we present more qualitative results of our method for 1-way 1-shot segmentation.

\noindent
\textbf{Foreground similarity maps.}
Since similarity maps between features and foreground prototypes provide an intuitive illustration of the model's semantic discriminative ability, we visualize more corresponding heatmaps in \cref{fig:more_simmap}. The results in the third row indicate that the baseline achieves high semantic discrimination for base classes in the source domain, but the features of the entire class are overly homogenized, failing to effectively distinguish finer-grained internal parts. When applied to target domains, the baseline exhibits insufficient discrimination for novel classes with varying segmentation granularities, \ie, the model suffers from semantic over-alignment. In contrast, results in the fourth row demonstrate that our method extracts hierarchical semantic features, thereby effectively adapting to novel classes.

\begin{figure}[t!]
\centering
    \includegraphics[width=1.0\linewidth]{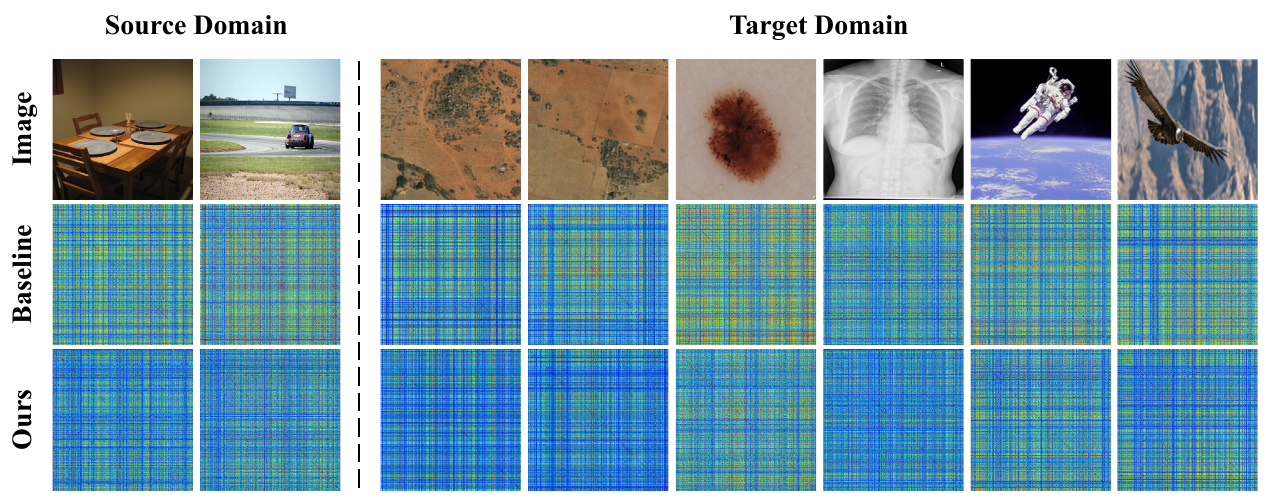}
    \caption{The channel-wise correlation maps show that our method mitigates attribute over-alignment and reduces channel redundancy. Brighter colors represent higher correlation values.}
    \label{fig:more_corr}
\end{figure}

\noindent
\textbf{Channel-wise correlation maps.}
We visualize channel-wise correlation maps of features, as shown in \cref{fig:more_corr}. The baseline exhibits higher channel correlations in both source and target domains, indicating that attribute over-alignment occurs during training. This causes channels encoding source-insensitive attributes to gradually degrade, with most channels tending to encode source-sensitive attributes jointly. In contrast, our method alleviates channel misalignment through hierarchical channel aggregation, thereby preserving more diverse attributes to adapt to different target domains.

\section{Limitations and Future Work}
\label{sec:limitation}
Similar to prototype-based methods, our method performs direct support-query matching using features extracted by the encoder. This paradigm exhibits high robustness in cross-domain scenarios and accurately localizes foreground classes, but yields limited segmentation precision at boundaries and other fine details. In contrast, decoder-based methods achieve higher boundary accuracy but are more prone to overfitting in cross-domain scenarios, failing to correctly localize foreground classes. Effectively combining the strengths of both paradigms represents a promising direction for future research.

Furthermore, although Vision Foundation Models (VFM) have been widely adopted in standard FSS tasks, their exploration in CD-FSS remains limited. Effectively leveraging the rich priors of VFMs to address CD-FSS is a valuable future direction.

%
%
\bibliographystyle{splncs04}
\bibliography{main}